\documentclass{new_tlp}
\usepackage{mathptmx}

\usepackage{xspace}

\usepackage{thm-restate}
\declaretheorem[numberwithin=section]{definition}

\usepackage{amsmath,amssymb}
\usepackage{graphicx}
\usepackage{caption}
\usepackage{subcaption}
\usepackage{todonotes}
\usepackage{xspace}
\usepackage{url}
\usepackage[hidelinks]{hyperref}
\usepackage{booktabs}
\usepackage{multirow}
\usepackage{marvosym}
\usepackage{tikz}
\usepackage{pgfplots}

\newtheorem{exmp}{Example}[section]

\usepackage{listings}
\lstdefinestyle{asp-style}{
	language=Prolog,
	keywordstyle=\linespread{1}\footnotesize\ttfamily,
	basicstyle=\linespread{1}\footnotesize\ttfamily,
	breaklines=true,
}
\lstdefinestyle{asp-style2}{
	language=Prolog,
	keywordstyle=\linespread{1}\footnotesize\ttfamily,
	basicstyle=\linespread{1}\footnotesize\ttfamily,
	breaklines=true,
}

\newcommand{\aspcore}{$\mathsf{ASP\text{-}Core\text{-}2}$\xspace}
\newcommand{\derives}{\mbox{\,:\hspace{0.1em}\texttt{-}}\,\xspace}

\newcommand{\Or}{\ensuremath{\ |\ }\xspace}
\newcommand{\p}{\ensuremath{{P}}\xspace}
\newcommand{\GP}{\emph{Ground(\p)}}
\newcommand{\naf}{\ensuremath{not\ }\xspace}

\def\Facts{\emph{Facts}}
\def\EDB{\emph{EDB}}
\def\IDB{\emph{IDB}}
\newcommand{\BP}{\ensuremath{B_{\p}}\xspace}
\newcommand{\UP}{\ensuremath{U_{\p}}\xspace}

\newcommand{\dlv}{\textsc{dlv}\xspace}
\newcommand{\idlv}{\textsc{i-dlv}\xspace}
\newcommand{\dlvdue}{\textsc{dlv\small{2}}\xspace}
\newcommand{\dlvdueserver}{\textsc{dlv}{\small{2}}-\textsc{server}\xspace}
\newcommand{\owltodlv}{\textsc{owl}{\small{2}}\textsc{dlv}\xspace}

\title[Precomputing Datalog evaluation plans in large-scale scenarios]{Precomputing Datalog evaluation plans\\ in large-scale scenarios
}
\author[Fiorentino et al.]{
Alessio Fiorentino, 
Nicola Leone, 
Marco Manna, 
Simona Perri, 
Jessica Zangari 
\\
Department of Mathematics and Computer Science, University of Calabria, Rende, Italy \\
\email{{\em lastname}@mat.unical.it} - \url{https://www.mat.unical.it}
}
\jdate{}
\pubyear{}
\submitted{-}
\revised{-}
\accepted{-}

\begin{document}

\label{firstpage}

\maketitle              
\begin{abstract}
With the more and more growing demand for semantic Web services over large databases, an efficient evaluation of Datalog queries is arousing a renewed interest among researchers and industry experts. In this scenario, to reduce memory consumption and possibly optimize execution times, the paper proposes novel techniques to determine an optimal indexing schema for the underlying database together with suitable body-orderings for the Datalog rules. The new approach is compared with the standard execution plans implemented in DLV over widely used ontological benchmarks. The results confirm that the memory usage can be significantly reduced without paying any cost in efficiency. This paper is under consideration in Theory and Practice of Logic Programming (TPLP).
\end{abstract}

\begin{keywords}
Datalog;
Query Answering;
Ontologies;
Query-plan;
Data Indexing
\end{keywords}

\section{Introduction}
Ontological reasoning services represent fundamental features in the development of the Semantic Web. Among them, scientists are focusing their attention on the so-called {\em ontology-based query answering} (OBQA), where a Boolean query has to be evaluated against a {\em logical theory} ({\em knowledge base}) consisting of an extensional {\em database} paired with an {\em ontology}~\cite{DBLP:conf/dlog/CaliGL09,DBLP:conf/amw/Ortiz13,DBLP:conf/ijcai/AmendolaLM18}.
%
A number of effective practical approaches proposed in the literature rewrite the query and the ontology into an equivalent Datalog program
~\cite{DBLP:conf/kr/CarralDK18,DBLP:conf/aaai/EiterOSTX12,DBLP:conf/ijcai/KontchakovLTWZ11,DBLP:conf/dlog/StefanoniMH12,DBLP:conf/ijcai/XiaoCKLPRZ18}.

With the more and more growing availability of large databases, however, an efficient yet memory-saving evaluation of Datalog queries is arousing a renewed interest among researchers and industry experts.
Typically, classical Datalog reasoners adopt sophisticated internal policies to speed-up the computation trying to limit the memory consumption. However,  when the amount of data exceeds a certain size, these policies may result inadequate. This happens, for instance, for
the full-fledged Datalog system \idlv~\cite{DBLP:journals/ia/CalimeriFPZ17,calimeri_perri_zangari} --- originally conceived as grounding engine in \dlvdue~\cite{DBLP:conf/lpnmr/AlvianoCDFLPRVZ17}.
Recently, to cope with large-scale scenarios, \idlv has been further optimized and partially re-engineered
by implementing novel techniques and heuristics to reduce memory consumption and possibly optimize execution times.
This process gave rise to even two branches of the system called \dlvdueserver~\cite{dlv2server} and \owltodlv~\cite{owl2dlv}.
In this paper, we present and evaluate one of the key approaches that is at the basis of the aforementioned improvements: the precomputation via Answer Set Programming (ASP)~\cite{gelf-lifs-91} of an ``evaluation plan'' for a given Datalog program.

To understand the principles underlying the new technique, let us first recall that  \idlv historically uses strategies for join orderings and indexing that are applied rule-by-rule at runtime and that are based on local statistics over data that become available during the computation.
%
As a result, for databases up to a few millions tuples, these stategies ensure fast evaluation at the expense of a reasonable amount of extra memory. Conversely, for databases with billions of tuples, both the time and the space used for implementing these strategies are too high.
To regain usability, the idea is to precompute a global 
indexing schema for the underlying database associated with suitable body-orderings for all the program rules.
On the one hand, this approach is less informed since, being implemented as a preprocessing phase, it cannot rely on any relevant information known during the computation, possibly leading to worsening in time. On the other hand, this allows to save both the time and space needed for computing/storing this information. Moreover, the global view on the program allows for a more parsimonious choice of the indices.
To make up for the lack of local statistics, our approach is based on the natural assumption that, when dealing with  very large databases, some information and statistics about the user domain are known in advance since they do not vary as fast as the actual data. This is the case, for example, for primary keys, foreign keys, small relations or selectivity of attributes.
Our contribution can be therefore summarized as follows:

\begin{itemize}
    \item Given a Datalog program $\mathcal{P}$, a database $\mathcal{D}$ and some domain properties, we define the notion of {\em evaluation plan},
    which consists of an indexing schema for $\mathcal{D}$ together with a suitable body-ordering for each rule of $\mathcal{P}$. Moreover, to target  ``optimal'' plans among all  admissible ones, we identify a number of additional options, the combination of which induces different preference orderings among all plans.

    \item We encode the problem of finding an optimal Datalog evaluation plan in ASP, by making use of choice-rules, strong constraints, weak constraints, aggregates and negation.

    \item We implement optimal plans by adding annotations~\cite{DBLP:journals/ia/CalimeriFPZ17} to
    the original Datalog program $\mathcal{P}$. The annotated program will be
    the actual input for \idlv. In this way, \idlv  execution is forced to follow the plan without the need for any internal change to the system.  Nonetheless, optimal plans are sufficiently general to be implemented natively also in different Datalog engines that do not benefit from features like annotations.


    \item We design a well-behaved setting in the context of ontological reasoning with the aim of minimizing the memory consumption without paying in efficiency.

    \item We conduct an experimental evaluation over popular ontological benchmarks widely used for testing both capabilities and performance of OBQA systems. In particular, we compare performance in terms of time and memory usage of \dlv when the classical computation is performed, and when the computation is driven by the planner. The results confirm that our plans improve the computation with a general gain in both time and space.
\end{itemize}

As a final remark, in case of reasoners with a server-like behavior, such as \dlvdueserver and \owltodlv, evaluation plans play an extremely important role, and the advantage of precomputing an evaluation plan is even more evident.
Indeed, when the ontology is known in advance, it is possible to determine ``offline''
the optimal plan, and therefore further improve the reasoning phase with respect to both time and memory.

The present work builds on top of the extended abstract presented at JELIA 2019~\cite{DBLP:conf/jelia/AlloccaCFPZ19}. In particular,
apart from providing a comprehensive description of the approach,
in this paper we also enrich the planner from both the formal and practical side. Indeed, we improve the notion of admissible ordering,
we introduce and implement the notion of preferences, we introduce extra preferences, we identify and exploit extra info in the ontological context, and we enriched the experimental evaluation.

\medskip

In the following, after recalling ASP syntax and semantics, we formalize the concept of evaluation plan for Datalog and we illustrate the modelling of such plans via ASP; eventually, we report about our experiments before drawing some conclusions.

\section{Preliminaries}\label{subsec:ASP}
The standard input language for ASP systems is referred to as \aspcore~\cite{calimeri2012asp}. For the sake of simplicity, we focus next on the basic aspects of the language; for a complete reference to the \aspcore standard, and further details about advanced ASP features, we refer the reader to~\cite{calimeri2012asp} and the vast literature.

\medskip

A {\em term} is either a constant or a variable. If $t_1,\dots, t_k$ are terms and $p$ is a \emph{predicate symbol} of arity $k$, then $p(t_1,\dots, t_k)$ is an {\em atom}.
A {\em literal} $\,l$ is of the form $a$ or $\naf a$, where $a$ is an atom; in the former case $l$ is {\em positive}, otherwise {\em negative}.
A {\em rule $r$} is of the form\ \ $ \alpha_1\Or\cdots\Or\alpha_k \derives $ $ \beta_1, \dots, \beta_n,$ $\naf\, \beta_{n+1},\dots,\naf\, \beta_{m}. $\ \ where $m\geq 0$, $k \geq 0$; $\alpha_1,\ldots,\alpha_k$ and $\beta_1, \dots, \beta_{m}$ are atoms. We define $H(r)$ = $\{ \alpha_1,$ $\ldots,$ $\alpha_k\}$ (the {\em head} of $r$) and $B(r) = B^+(r) \cup B^-(r)$ (the {\em body} of $r$), where $B^+(r) = \{ \beta_1,$ $\dots,$ $\beta_n\}$ (the {\em positive body}) and $B^-(r)$ = $\{\naf\ \beta_{n+1},$ $\dots,$ $\naf\ \beta_{m} \}$ (the {\em negative body}).
If $H(r) = \emptyset$ then $r$ is a {\em (strong) constraint}; if $B(r) =\emptyset$ and $|H(r)| =1$ then $r$ is a {\em fact}.
A rule $r$ is safe if each variable of $r$ has an occurrence in $B^+(r)$\footnote{We remark that this definition of safety is specific for the syntax considered herein.
For a complete definition we refer the reader to~\cite{calimeri2012asp}.}.
An ASP program is a finite set $P$ of safe rules.
A program (a rule, a literal) is {\em ground} if it contains no variables.
In the following, a Datalog program is referred to as a finite set $P$ of safe rules stratified with respect to negation and without disjunction in the heads.
A program (a rule, a literal) is {\em ground} if it contains no variables.
A predicate is defined by a rule $r$ if it occurs in $H(r)$. A predicate defined only by facts is an \EDB\ predicate, the remaining are \IDB\ predicates. The set of all facts in $P$ is denoted by \Facts($P$); the set of instances of all \EDB\ predicates in $P$ is denoted by \EDB($P$).

Given a program $P$, the {\em Herbrand universe} of $P$, denoted by \UP, consists of all ground terms that can be built combining constants and function symbols appearing in $P$.
The {\em Herbrand base} of~$P$, denoted by \BP, is the set of all ground atoms obtainable from the atoms of $P$ by replacing variables with elements from \UP. A {\em substitution} for a rule $r \in P$ is a mapping from the set of variables of $r$ to the set \UP of ground terms.
A {\em ground instance} of a rule $r$ is obtained applying a substitution to $r$. The {\em full instantiation} \GP\ of $P$ is defined as the set of all ground instances of its rules over \UP. An {\em interpretation} $I$ for $P$ is a subset of \BP. A positive literal $a$ (resp., a negative literal $\naf\ a$) is true w.r.t. $I$ if $a \in I$ (resp., $a \notin I$); it is false otherwise. Given a ground rule $r$, we say that $r$ is satisfied w.r.t. $I$ if some atom appearing in $H(r)$ is true w.r.t. $I$ or some literal
appearing in $B(r)$ is false w.r.t. $I$. Given a program $P$, we say that $I$ is a {\em model} \ of $P$, iff all rules in \GP\ are satisfied w.r.t. $I$. A model $M$ is {\em minimal} if there is no model $N$ for $P$ such that $N \subset M$. The {\em Gelfond-Lifschitz reduct}~\cite{DBLP:journals/ngc/GelfondL91} of $P$, w.r.t.\ an interpretation $I$, is the positive ground program $P^I$ obtained from $\GP$ by: $(i)$ deleting all rules having a negative literal false w.r.t. $I$; $(ii)$ deleting all negative literals from the remaining rules. $I\subseteq \BP$ is an {\em answer set} for a program $P$ iff
$I$ is a minimal model for $P^I$.
The set of all answer sets for $P$ is denoted by $AS(P)$.

\section{Datalog Evaluation Plans}
In its default computational process, for optimizing the evaluation of each rule, \idlv determines body orderings and indices on demand, according to strategies taking into account only local information for the rule at hand. In more detail, before instantiating some Datalog rule, \idlv reorders the body literals  on the basis of some join-ordering heuristics~\cite{DBLP:journals/ia/CalimeriFPZ17}; then, according to the chosen ordering, it determines and creates needed indices. However, when memory consumption must be limited, an approach based on a global view over all rules, allowing for a more parsimonious creation of indices, is preferable.

In this section, we describe our approach for computing \emph{optimal evaluation plans} for a set $\mathcal{P}$ of positive Datalog rules to be evaluated over an extensional database $\mathcal{D}$. We define an evaluation plan of $\mathcal{P}$ as an indexing schema over $\mathcal{P}\cup\mathcal{D}$ together with a suitable body-ordering for each rule of $\mathcal{P}$. An indexing schema consists of the set of indices adopted to instantiate all rules in $\mathcal{P}$ over $\mathcal{D}$.
Our approach makes use of an ASP program for computing an \emph{optimal} evaluation plan $\mathcal{E}$ of $\mathcal{P}$ in a preprocessing phase; then $\mathcal{P}$ is annotated with directions that force \dlv computation to follow $\mathcal{E}$ when evaluating $\mathcal{P}$.

In the following, after a formal definition of an \emph{evaluation plan}, we introduce the notion of \emph{strategy} and then we specify the concept of \emph{optimal} evaluation plan w.r.t. a certain strategy.

\subsection{Admissible Plans}\label{subsec::evaluationplans}
Let $\mathcal{P}$ be a set of positive Datalog rules with non-empty body and let $\mathcal{D}$ be a database, i.e. a set of facts. We indicate with pred$(\mathcal{P}\cup\mathcal{D})$ the set of all predicates occurring in $\mathcal{P}\cup\mathcal{D}$ and with rel($p$) the set $\{\alpha \in \mathcal{D} : pred(\alpha) = p\}$ of the elements of $\mathcal{D}$ sharing the predicate name $p$. We write $p[i]$ to indicate the $i$-th argument of the predicate $p$.
In the following, after  formalizing the standard notions of \emph{ordering} of a rule and \emph{indexing schema} of a Datalog program, we introduce the novel notion of  \emph{evaluation plan}
along with some preliminary definitions.

\begin{definition}
Let $r$ be a rule in $\mathcal{P}$ and $B(r)$ be the set of the atoms appearing in the body of $r$. Let $F_a$ be a (possibly empty) subset of atoms in $B(r)$ and $F_p$ be a subset of $\{1,\cdots,|B(r)|\}$.  A \emph{position assignment on $r$} is a one-to-one map $p_r:F_a\rightarrow F_p$. A pair $(\alpha,p)$ such that $p_r(\alpha)=p$ is called a \emph{fixed position} w.r.t. $p_r$. An \emph{ordering on $r$} is a bijective function $pos(r,\cdot):B(r)\rightarrow\{1,\cdots,|B(r)|\}$. Having fixed a position assignment $p_r$ on $r$, we define a \emph{$p_r$-ordering on $r$} as an ordering on $r$ such that $pos(r,\alpha)=p_r(\alpha)$ for each $\alpha\in F_a$.
\end{definition}

The definition above presents a body ordering as a rearrangement of the literals in the body, but notably, allows for having a certain number of atoms in the body in some fixed positions. This is because, according to the knowledge of the domain at hand, if one is aware that a particular choice for the orderings is convenient, the planner can be driven so that only plans complying with this choice are identified.

\begin{definition}
Let $U:=\{p[i] : p\in\textrm{pred}(\mathcal{P}\cup\mathcal{D}),\,1 \leq i \leq a(p)\}$, where $a(p)$ represents the arity of the predicate $p$. An \emph{indexing schema} $\mathcal{S}$ over $\mathcal{P}\cup\mathcal{D}$ is a subset of $U$. Given a subset $I\subseteq U$, we say that $\mathcal{S}$ fixes $I$ if $I\subseteq \mathcal{S}$.
\end{definition}

Intuitively, an indexing schema is a subset of the arguments of all predicates in pred$(\mathcal{P}\cup\mathcal{D})$. Furthermore, similarly to the definition of ordering that may allow for fixed positions, we give the possibility to fix also a set of indices.

\begin{exmp}\label{running_example}
As a running example in this section, consider the following positive Datalog rule $r:$
\begin{lstlisting}[style=asp-style2]
    h(X,Z,W) :- a(X,Z), b(V,W), c(Z), d(V), e(Y,Z).
\end{lstlisting}
Let's consider the position assignment $p_r$ which fixes the atom $b(V,W)$ in first position.
A possible $p_r$-ordering may be:
\begin{align*}
    pos(r,a(X,Z))=3,\hspace{0.5 em}&pos(r,b(V,W))=1,\hspace{0.5 em}pos(r,c(Z))=5,\\
    pos(r,d(V))=2,\hspace{0.5 em}&pos(r,e(Y,Z))=4.
\end{align*}
By means of such ordering the body atoms of $r$ are rearranged as follows:
\begin{lstlisting}[style=asp-style2]
    h(X,Z,W) :- b(V,W), d(V), a(X,Z), e(Y,Z), c(Z).
\end{lstlisting}
The set $\mathcal{S}:=\{a[2],c[1],d[1],e[2]\}$ is an example of indexing schema over the predicates appearing in $r$.
\end{exmp}

\medskip

With a rule $r\in\mathcal{P}$ we can associate a hypergraph $H(r) = (V,E)$ whose vertex set $V$ is the set of all terms appearing in $B(r)$ and the edges in $E$ are the term sets of each atom in $B(r)$. Given a rule $r$ of $\mathcal{P}$, a \emph{connected component} of $r$ is a set of atoms in $B(r)$ that define a connected component in $H(r)$.

\begin{exmp}

Let $r$ be the rule of Example~\ref{running_example}. The hypergraph $H(r)$ associated to $r$ has $V=\{ X,Y,Z,V,W\}$ and $E=\{\{ X,Z\},\{V,W\},\{Z\},\{V\},\{Y,Z\}\}.$
The connected components of $r$ are $C_1=\{a(X,Z),\\c(Z),e(Y,Z)\}$ and $C_2=\{b(V,W),d(V)\}$.

\end{exmp}




Let us introduce now the notions of \emph{separation} between two connected components and \emph{well ordering} of a component of a rule.
\begin{definition}\label{separated_components}
Let $r$ be a rule of $\mathcal{P}$ and $pos(r,\cdot)$ be an ordering on $r$. Two connected components $C_1$ and $C_2$ of $r$ are \emph{separated} w.r.t. $pos(r,\cdot)$ if $\max\{pos(r,\alpha):\alpha\in C_1\}<\min\{pos(r,\beta):\beta\in C_2\}$ or vice versa.
\end{definition}

An argument of an atom appearing in the body of a rule $r$, is said to be \emph{bound}, w.r.t. an ordering on $r$, if it is either a constant or a variable appearing in a previous atom, and is said to be \emph{indexBound}, w.r.t. an ordering on $r$ and an indexing schema $\cal S$, if it is bound and it belongs to the schema $\mathcal{S}$. The definition below provides the notion of \emph{well ordering} of a connected component in a rule.
\begin{definition}\label{well-ordering}
Let $r$ be a rule of $\mathcal{P}$, $\mathcal{S}$ be an indexing schema and $pos(r,\cdot)$ be an ordering on $r$. A connected component $C$ of $r$ is \emph{well-ordered w.r.t. $\mathcal{S}$ and $pos(r,\cdot)$} if, assuming $m=\min\{pos(r,\alpha):\alpha\in C\}$, for each $\beta\in C$ with $pos(r,\beta)=j$ and $j>m$, it holds that: $(i)$ $\beta$ has at least an argument  which is indexBound, and $(ii)$ either all the arguments of $\beta$ are bound or there is no other atom in a later position (in the same component) that, placed in place of $\beta$, would have all the arguments bound.
\end{definition}
\begin{exmp}
Let's consider the rule $r$ with the ordering $pos(r,\cdot)$ and the indexing schema $\cal S$ as in our running example.
The two connected components of $r$ are clearly separated w.r.t. $pos(r,\cdot)$. The indexBound arguments w.r.t. $pos(r,\cdot)$ and $\cal S$ are $c[1]$, $d[1]$ and $e[2]$. It can be easily seen that the connected component $C_2$ is well-ordered w.r.t. $\cal S$ and $pos(r,\cdot)$. The same cannot be said for the component $C_1$; in fact, not all the arguments of the atom $e(Y,Z)$ are bound and the atom $c(Z)$, positioned in place of $e(Y,Z)$, would have all the arguments bound.
\end{exmp}

The notion of separation among connected components is needed for identifying, within rule bodies, clusters of literals that do not share variables. The idea is  that the ordering computed by the planner should keep separated these clusters in order to avoid, as much as possible, the computation of Cartesian products during the instantiation; at the same time, literals within the clusters are properly rearranged in order to comply with the selected indexing schema, thus avoiding the creation of further indices.

Next, we provide the \emph{admissibility} property which, in turn, characterizes the evaluation plans.

\begin{definition}
Given a rule $r\in\mathcal{P}$ and an indexing schema $\mathcal{S}$, we say that an ordering $pos(r,\cdot)$ is \emph{admissible} w.r.t. $\mathcal{S}$ if the connected components of $r$ are mutually separated (w.r.t. $pos(r,\cdot)$) and well-ordered (w.r.t. $pos(r,\cdot)$ and $\mathcal{S}$).
\end{definition}


We define below an evaluation plan for a Datalog program.
\begin{definition}
Let (i) $\{p_r\,;\,\, r \in \mathcal{P}\}$ be a given set of position assignments, and (ii) $I$ be a given subset of $\{p[i] : p\in\textrm{pred}(\mathcal{P}\cup\mathcal{D}),\,1 \leq i \leq a(p)\}$. An \emph{evaluation plan} $\mathcal{E}$ of $\mathcal{P}$ consists of an indexing schema $\mathcal{S}$ that fixes $I$ together with a $p_r$-orderings for each $r\in\mathcal{P}$ being admissible w.r.t. $\mathcal{S}$.
We say that $\mathcal{P}$ enjoys an \emph{efficient evaluation} if it is associated to an evaluation plan.
\end{definition}


\begin{exmp}
In our running example, the ordering $ord(r,\cdot)$ is not admissible w.r.t. the schema $\cal S$. Thus, $\cal S$ and $ord(r,\cdot)$ do not represent an evaluation plan of the program ${\cal P}=\{r\}$. However, an evaluation plan of $\cal P$ could be obtained by exchanging the assignments of the atoms $c(Z)$ and $e(Y,Z)$ in $ord(r,\cdot)$. It would be appropriate to note that, in the latter case, we would obtain a further evaluation plan by excluding the argument $a[2]$ from the indexing schema (and thus saving an index). Starting from this consideration, we introduce the concepts of preference and evaluation strategy in the next section.
\end{exmp}

\subsection{Preferences}\label{preferences}

Let $\cal P$ be a positive Datalog program, $E_{\cal P}$ be the set of all the evaluation plans of $\cal P$ and $w:E_{\cal P}\rightarrow \mathbb{N}$ be a function that we call \emph{cost function on $E_{\cal P}$}.
Given two evaluation plans $\mathcal{E}_1,\mathcal{E}_2\in E_\mathcal{P}$, we say that \emph{$\mathcal{E}_1$ is preferable to $\mathcal{E}_2$ w.r.t. the function cost $w$} if $w(\mathcal{E}_1) < w(\mathcal{E}_2)$, while we say that \emph{$\mathcal{E}_1$ is equivalent to $\mathcal{E}_2$ w.r.t.  $w$} if $w(\mathcal{E}_1) = w(\mathcal{E}_2)$. Moreover, consider a finite set $W = \{w_1,\dots,w_n\}$ of cost functions on $E_{\cal P}$, we define an \emph{evaluation strategy for $\cal P$} as a finite sequence $\Sigma = (w_{\delta_1},\dots,w_{\delta_k})$ of distinct elements in $W$. We say that \emph{$\mathcal{E}_1\in E_\mathcal{P}$ is preferable to $\mathcal{E}_2\in E_\mathcal{P}$ w.r.t. the strategy $\Sigma=(w_1,\dots,w_k)$} if either:
\begin{itemize}
    \item $\mathcal{E}_1$ is preferable to $\mathcal{E}_2$ w.r.t. $w_1$, or
    \item there exists $j\in\{2,\dots,k\}$ such that $\mathcal{E}_1$ is equivalent to $\mathcal{E}_2$ w.r.t.  $w_i$ for each $i=1,\dots,j-1$, and $\mathcal{E}_1$ is preferable to $\mathcal{E}_2$ w.r.t. $w_j$.
\end{itemize}
According to the notion of \emph{preference} of an evaluation plan over another w.r.t. a strategy, we can now introduce the definition of ``optimal'' plans w.r.t. that strategy.
\begin{definition}
Let $E_{\cal P}$ be the set of all the evaluation plans for a positive Datalog program $\cal P$ and $\Sigma$ be an evaluation strategy for $\cal P$. An evaluation plan ${\cal E}_0$ is said to be \emph{optimal w.r.t. $\Sigma$} if it is either preferable or equivalent to each ${\cal E} \in E_{\cal P}$ w.r.t. $\Sigma$.
\end{definition}
Intuitively, finding the optimal evaluation plans against a strategy $\Sigma = (w_1,\dots,w_k)$ means finding those that minimize the cost function $w_1$, then, among these, find those that minimize the function $w_2$, and so on. We report next four functions used for defining our evaluation strategies.
From now on when we talk about the  functions $w_1$, $w_2$, $w_3$ and $w_4$ we will refer to the following:
\begin{itemize}
    \item $w_1(\mathcal{E}):=\sum_{p[i]\in\mathcal{S}}c(p,i)$, where $c(p,i)$ is the cost of building an index over $p[i]$ in the indexing schema $\cal S$. Note that we presuppose the knowledge of $c(p,i)$ values. Such costs can be estimated via heuristics or actually computed, depending on the application domain at hand.
    As said in the introduction, the novel approach is based on the natural assumption that, when dealing with very large databases, some information and statistics about the user domain are known in advance since they do not vary as fast as the actual data. Apart from primary keys and foreign keys, which in OBQA are related to the ontological schema, some statistics on the data can be also taken into account. This is the case, for example, of the {\em estimation of the selectivity} of an attribute, which gives an indication of the average number of times that a constant (or individual) occurs in the relation in correspondence of the given attribute (note that, the special case of estimation of the selection equal to $1$ indicates that the given attribute is actually a key). This value can be taken into account for estimating the size (and thus, the cost) of an index for the given attribute.

    \item $w_2(\mathcal{E})$ is defined as the sum of the positions of atoms involved in recursion.
    We prefer that atoms involved in recursion are placed as soon as possible. The extension of such atoms might considerably grow and change during computation; placing them before other atoms in the body could avoid the creation of expensive indices.
    \item $w_3(\mathcal{E})$ is the number of indices set on arguments that are not primary keys. In other words we prefer indices set on arguments representing primary keys.
    \item $w_4(\mathcal{E}):=\sum_{r\in\mathcal{P}}\sum_{\alpha\in B(r)}[maxArity-u(\alpha,r)]*pos(r,\alpha)$, where $maxArity$ represents the maximum arity of the atoms appearing in $\cal P$ and $u(\alpha,r)$ is the number of unbound arguments of the atom $\alpha$ in the rule $r$.
    We prefer that atoms having large number of unbound arguments (that is, those that minimize the first factor in the above summation) are placed as soon as possible as they possibly will lead to have new completely bound atoms to be placed in successive positions.
\end{itemize}

\section{ASP-based Implementation}


In the following we describe the ASP code devised in order to compute optimal evaluation plans. For the sake of simplicity, as the program is rather long and involved, we report here only some key parts; the full ASP code is available online.%
\footnote{See \url{https://www.mat.unical.it/perri/iclp2019.zip}.}

The program is based on the classical ``Guess/Check/Optimize'' paradigm  and combines: (i) \textit{choice and disjunctive rules} to guess an indexing schema $\mathcal{S}$ over $\mathcal{P}\cup\mathcal{D}$ and, for each rule $r$ in $\mathcal{P}$, an ordering ord$(r,\cdot)$; (ii) \textit{strong constraints} to guarantee, for each rule $r$, the admissibility of ord$(r,\cdot)$ w.r.t. $\mathcal{S}$; (iii) \textit{weak constraints} to find out the optimal evaluation plans of $\mathcal{P}$ w.r.t. the chosen strategy.

\subsection{Data Model}

The planner consists of an ASP program taking as input a set of facts representing $\mathcal{P}$ and the database $\mathcal{D}$; each rule of $\mathcal{P}$ is represented by means of facts of the form:

\begin{lstlisting}[style=asp-style]
    rule(Rule,Description,NumberOfBodyAtoms).
    headAtom(Rule,Atom,Predicate).
    bodyAtom(Rule,Atom,Predicate).
    sameVariable(Rule,Atom1,Arg1,Atom2,Arg2).
    constant(Rule,Atom,Arg).
\end{lstlisting}
Facts over the predicate ${\tt rule}$ associate each rule $r$ to an identifier and provide the number of its body atoms. Atoms in the body and in the head of each rule $r$ are represented by ${\tt bodyAtom}$ and ${\tt headAtom}$ predicates respectively. The predicate ${\tt sameVariable}$ provides the common variables related to every pair of atoms appearing in $r$, whereas ${\tt constant}$ states that a constant term occurs in the argument of an atom of $r$. An example of the basic input concepts described above is the following:
\begin{exmp}
The following program $\cal P:$
\begin{lstlisting}[style=asp-style]
    h1(X) :- a(X,Y),b(Y).
    h2(Y) :- a(Y,X).
\end{lstlisting}
is represented by means of the facts:
\begin{lstlisting}[style=asp-style]
    rule(0,"h1(X):-a(X,Y),b(Y).",2).
    headAtom(0,"h1(X)","h1/1").
    bodyAtom(0,"a(X,Y)","a/2").
    bodyAtom(0,"b(Y)","b/1").
    sameVariable(0,"h1(X)",1,"a(X,Y)",1).
    sameVariable(0,"a(X,Y)",2,"b(Y)",1).

    rule(1,"h2(Y):-a(Y,X).",1).
    headAtom(1,"h2(Y)","h2/1").
    bodyAtom(1,"a(Y,X)","a/2").
    sameVariable(1,"h2(Y)",1,"a(Y,X)",1).
\end{lstlisting}
\end{exmp}

The database $\mathcal{D}$ is represented by means of facts over predicate {\tt relation}, while the costs of building indices over arguments are given by facts over predicate $\tt index$.
\begin{lstlisting}[style=asp-style]
    relation(Predicate,Arity).
    index(Predicate,Arg,Cost).
\end{lstlisting}

Furthermore, the planner allows for having a certain number of atoms in the body in some previously fixed positions and a set of indices fixed in $\cal S$. This is because, according to the knowledge of the domain at hand, if one is aware that a particular choice for the orderings and the indexing policy is convenient, the planner can be driven so that only plans complying with this choice are identified. The planner can also exploit the presence of arguments representing primary keys for predicates in $\mathcal{P} \cup \mathcal{D}$. Such information, if available, can be given in input to the ASP planner by means of facts of the form:

\begin{lstlisting}[style=asp-style]
    fixedPosition(Rule,Atom,Pos).
    fixedIndex(Predicate,Arg).
    key(Predicate,Arg).
\end{lstlisting}


\subsection{Guess Part}

The following choice rule~\cite{calimeri2012asp} guesses a subset of the arguments of all predicates, namely an indexing schema $\mathcal{S}$, over $\mathcal{P}\cup\mathcal{D}$. Notably, the arguments to be indexed are chosen among a restricted set of arguments, called \textit{indexable}, in order to keep the search space smaller. For instance, arguments that are not involved in joins are not indexable.

\begin{lstlisting}[style=asp-style]
    {setIndex(Predicate,Arg)} :- indexable(Predicate,Arg).
\end{lstlisting}

Beside this choice rule, the guess part contains also the following rule for guessing a body-ordering for each rule $r$ in $\mathcal{P}$. In particular, the choice guesses a position in the body for each atom whose position has not been previously fixed ($\tt fixedAtomRule$).
Clearly, only positions not already occupied by another body atom in the same rule ($\tt fixedPositionRule$) are guessable. The predicates $\tt fixedAtomRule$ and $\tt fixedPositionRule$ are computed according to the predicate $\tt fixedPosition$ described above.
 \begin{lstlisting}[style=asp-style]
    {pos(Atom,Rule,Pos):position(Pos),Pos>=1,Pos<=Size,
       not fixedPositionRule(Rule,Pos)}=1 :- rule(Rule,Size),
       bodyAtom(Rule,Atom,_),not fixedAtomRule(Rule,Atom).
\end{lstlisting}


\subsection{Check Part}
This part discards, by means of \textit{strong constraints}, solutions that do not satisfy (according to the definitions in Section~\ref{subsec::evaluationplans}) the conditions to be considered admissible evaluation plans.
In particular, conditions that have to be necessarily satisfied are the following:
\begin{enumerate}
    \item 
    The \emph{connected components} of each rule of $\cal P$ must be kept separate. According to the definition~\ref{separated_components}, this is ensured by the following constraint.

    \begin{lstlisting}[style=asp-style]
    :- pos(Atom1,Rule,Pos1),pos(Atom2,Rule,Pos2),
       sameComponent(Rule,Atom1,Atom2),pos(Atom3,Rule,Pos3),
       not sameComponent(Rule,Atom1,Atom3),Pos1<Pos3,Pos3<Pos2.
    \end{lstlisting}

    \item According to the first point of the definition~\ref{well-ordering},
    to guarantee that each connected component is \emph{well-ordered}, each atom, except those in the first position of each component, must have at least an argument \emph{indexBound}. This condition is guaranteed by the constraint below, where predicates $\tt firstPosition$ and $\tt indexBound$ suggest, respectively, the first positions of the components in each rule, and the indexBound arguments of each atom in a rule.

    \begin{lstlisting}[style=asp-style]
    :- pos(Atom,Rule,Pos),firstPosition(Rule,FirstPos),
       Pos>FirstPos,#count{Arg:indexBound(Arg,Atom,Rule)}=0.
    \end{lstlisting}

    \item The second condition of the definition~\ref{well-ordering} is modeled by means of the following constraint. Here the predicate $\tt atomVars$ indicates the number of variables occurring in every atom.

    \begin{lstlisting}[style=asp-style]
    :- pos(Atom,Rule,Pos),not boundAtom(Atom,Rule,Pos),
       pos(Atom1,Rule,Pos1),not boundAtom(Atom1,Rule,Pos1),
       Pos1>Pos,Pos2>Pos1,boundAtom(Atom2,Rule,Pos2),
       atomVars(Atom2,Rule,N),
       #count{Arg2:sameVariable(Rule,Atom2,Arg2,Atom,_)}=N.
    \end{lstlisting}



\end{enumerate}
The checking part contains also an additional constraint encoding the following basic check for guaranteeing the correctness of the plans. In particular, this basic check ensures that two different atoms do not occupy the same position in any rule:

    \begin{lstlisting}[style=asp-style]
    :- pos(Atom1,Rule,Pos),pos(Atom2,Rule,Pos),Atom1!=Atom2.
    \end{lstlisting}


\subsection{Optimize Part}
Eventually, in this section we describe the part for identifying the optimal evaluation plan according to the evaluation strategy that one decides to apply. Remember that a strategy is a finite combination of cost functions. Currently, the planner is equipped with the four cost functions described in  Section~\ref{preferences}, each of which is represented by a specific \textit{weak constraint}. Note that, weak constraints allow for expressing preferences possibly having different importance levels. The planner allows to fix these priority levels according to the chosen strategy by providing in input facts which indicate that the cost function $w_N$ has priority level $P$.
    \begin{lstlisting}[style=asp-style]
    priorityCostFunction(N,P).
    \end{lstlisting}

For instance, suppose we want to represent the strategy $\Sigma = (w_1,w_3,w_2)$, then we need the following input facts to indicate that the cost function $w_1$ has priority level $3$, $w_3$ has priority level $2$ and $w_2$ has priority level $1$. Note that in this case the cost function $w_4$ is not activated.
    \begin{lstlisting}[style=asp-style]
    priorityCostFunction(1,3).
    priorityCostFunction(3,2).
    priorityCostFunction(2,1).
    \end{lstlisting}

This means that the planner is customizable. Indeed, depending on the knowledge of the domain at hand, one can choose to adapt the strategy to his own needs simply by exchanging the priority levels of the cost functions among those already present in the planner, or even by integrating new cost functions (with the addition of new constraints in the encoding). In the following we illustrate the weak constraints representing the cost functions defined in Section~\ref{preferences}.
\begin{enumerate}
    \item 
    The rule below aims to minimize index occupation. To this end, we presuppose the knowledge of the costs (or their estimation) of building indices over arguments and we represent them by facts of form $\tt index(Predicate,Arg,Cost)$.

    \begin{lstlisting}[style=asp-style]
    :~ setIndex(Predicate,Arg),index(Predicate,Arg,Cost),
       priorityCostFunction(1,P). [Cost@P,Predicate,Arg,Cost]
    \end{lstlisting}

    \item We prefer that atoms involved in recursion are placed as soon as possible.
    The weak constraint makes uses of the auxiliary predicate $\mathtt{recursivePredicate}$ providing information about which predicates of the program are recursive. We do not report its definition for the sake of readability.

    \begin{lstlisting}[style=asp-style]
    :~ pos(Atom,Rule,Pos),bodyAtom(Rule,Atom,Predicate),
       recursivePredicate(Predicate),priorityCostFunction(2,P).
       [Pos@P,Rule,Pos]
    \end{lstlisting}

    \item Indices set on arguments representing primary keys are possibly preferred:

    \begin{lstlisting}[style=asp-style]
    :~ setIndex(Predicate,Arg),not key(Predicate,Arg),
       priorityCostFunction(3,P). [1@P,Predicate,Arg]
    \end{lstlisting}

    \item Atoms having large number of unbound arguments should be placed as soon as possible in the body. Also in this case we make use of an auxiliary predicate: $\mathtt{numBoundArgs}$ provides the number of bound arguments of an atom in a rule.

    \begin{lstlisting}[style=asp-style]
    :~ numBoundArgs(Atom,Rule,Pos,B),maxArity(N),
       bodyAtom(Rule,Atom,Predicate),relation(Predicate,Arity),
       priorityCostFunction(4,P). [(N-Arity+B)*Pos@P,Rule,Pos]
    \end{lstlisting}

\end{enumerate}


%


\section{Experimental Evaluation}

Hereafter we report the results of an experimental activity carried out to assess the effectiveness of the ASP-based evaluation planner.

\subsection{Benchmarks}

Our experimental analysis relies on four benchmarks: LUBM (Lehigh University BenchMark), LUBM-LUTZ, Stock Exchange and Vicodi.

\textbf{LUBM}. It is one of the most popular ontologies for testing both capabilities and performance of OBQA systems; indeed, it has been specifically developed to facilitate the evaluation of Semantic Web reasoners in a standard and systematic way. In particular, the benchmark is intended to evaluate performance of those reasoners with respect to extensional queries over large databases that refer to a single realistic ontology. The LUBM benchmark consists of a university domain OWL 2 ontology along with customizable and repeatable synthetic data and a set of 14 SPARQL queries\footnote{LUBM is available at \url{http://swat.cse.lehigh.edu/projects/lubm/}.}. Queries 2, 6, 9 and 14  involve constants, while the other queries are constant-free. In our experiments, the original LUBM ontology and the official 14 queries have been translated into Datalog via the {\sc clipper} system~\cite{DBLP:conf/aaai/EiterOSTX12}. The official LUBM generator has been adopted to generate four databases of increasing sizes: LUBM-500, LUBM-1,000, LUBM-2,000 and LUBM-4,000, where the number associated to each database name indicates the number of universities composing it. The number of facts in the databases ranges from about 67,000,000 to about half a billion facts.

\textbf{LUBM-LUTZ}. It is a variant of LUBM designed by~\citeN{DBLP:conf/semweb/LutzSTW13}. This benchmark consists of an OWL2 ontology and 11 queries (both different from those of LUBM) along with a modified version of the LUBM official generator allowing to set the level of incompleteness in the database. As done for LUBM, the ontology and the queries have been translated into Datalog via the {\sc clipper} system~\cite{DBLP:conf/aaai/EiterOSTX12}. All queries are without constants. We generated five databases of increasing sizes and having an incompleteness percentage of $10\%$: LUTZ-500, LUTZ-1,000, LUTZ-2,000, LUTZ-4,000 and LUTZ-8,000. Again, the number associated to each database name indicates the number of universities composing it.

\textbf{Stock Exchange and Vicodi}. These are two real world ontologies widely used in literature for the evaluation of query rewriting systems~\cite{DBLP:conf/semweb/MoraC13}. For each of these two ontologies, we selected $5$ queries featuring constants and we used the SyGENiA generator~\cite{DBLP:journals/jair/GrauMSH12} to produce five databases having from $1,000$ to $40,000$ tuples and a number of individuals varying from $100$ to $4,000$. These are the maximum sizes that can be generated using SyGENiA.

\subsection{Setting}

Experiments on LUBM-LUTZ have been performed on a Dell Linux server with an Intel Xeon Gold 6140 CPU composed of 8 physical CPUs clocked at 2.30 GHz, with 297GB of RAM. Experiments on LUBM, Vicodi and Stock Exchange have been performed on a NUMA Linux machine equipped with two 2.8 GHz AMD Opteron 6320 processors and 128GB RAM. Unlimited time and memory were granted to running processes. Benchmarks and executables used for the experiments are available at \url{https://www.mat.unical.it/perri/iclp2019.zip}.

Two different executions have been compared: $(i)$ a classical execution of \idlv which, given as input the so generated encodings, chooses body orderings and indexing strategies with its default policies, and $(ii)$ an execution driven by the planner in which \idlv is forced to follow the precomputed evaluation plan that decided body orderings and indices in order to reduce memory consumption. These constraints have been defined via annotations, that represent specific means to express preferences over its internal computational process~\cite{DBLP:journals/ia/CalimeriFPZ17}.

\subsection{Planner customization}\label{sec:custom}


In the context of OBQA, where the objective is to answer a query, the rewritten Datalog program typically benefits from the application of the so-called Magic Sets technique~\cite{DBLP:journals/ai/AlvianoFGL12}. This produces a new equivalent program containing extra intensional predicates that could have very small extensions during the computation. These predicates, in a setting where memory consumption should be limited, could be moved towards the end of the body so that it is more likely saving space for needed indices.  Hence, to instantiate our planner, we consider this additional domain information and use facts of the form {\tt fixedPosition(Rule,Atom,Pos)} to specify it. It is worth remarking that in \idlv this customization has an impact only in case of queries featuring constants since magic atoms are not generated for queries without constants.
Furthermore, for all attributes involved in extensional relations, we provide, via facts of the form {\tt index(Predicate,Arg,Cost)}, an estimation of the size of an index for that attribute. In particular,
in our experiments, to have available this information we generate and analyze a ``small'' database for each benchmark.

In our experiments, we considered different planner customizations depending on the domain at hand. In particular, for LUBM, Vicodi and Stock Exchange,  consisting mainly of queries featuring constants, we adopted the strategy $\Sigma_1 = (w_2, w_4)$. The idea underlying this choice is that, in such domains, \idlv can benefit from the Magic Sets technique and fixing the positions of magic atoms as described above is already sufficient to drive the planner. On LUBM-LUTZ, having instead constant-free queries, we adopted the strategy $\Sigma_2 = (w_1,w_2,w_3, w_4)$; indeed, since Magic Sets are not active, no fixed positions can be provided and a richer strategy is necessary for avoiding an almost blind plan computation.



\subsection{Discussion}
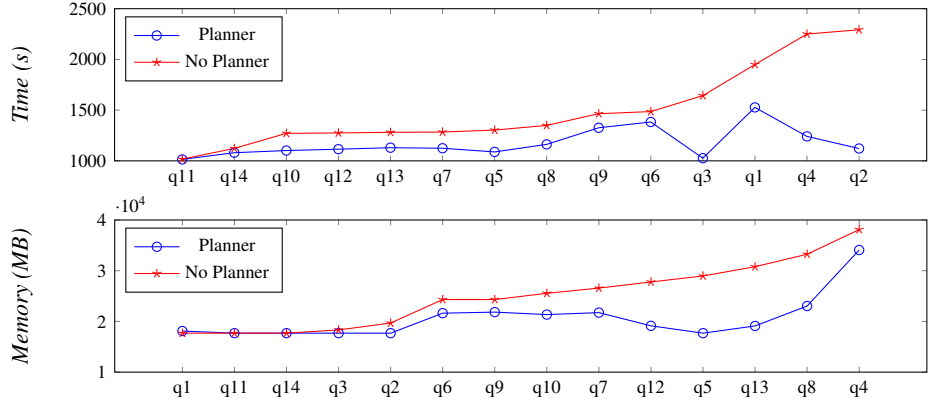
\begin{figure}[h]\centering
\centering
\pgfplotstableread[]{figures/timeLUBM.dat}\datatable
        \begin{tikzpicture}[scale=1]
    	\pgfkeys{%
    	    /pgf/number format/set thousands separator = {}}
    	\begin{axis}[
        	scale only axis
        	, font=\scriptsize
        	, ylabel={\textit{\small Time (s)}}
        	, width=0.8\textwidth
            , legend style={at={(0.11,0.95)},anchor=north}
          	, height=0.15\textwidth
        	, ymin=1000, ymax=2500
        	, major tick length=2pt
        	, xtick=data
            , xticklabels from table={\datatable}{label}
    	]
    	\addplot [mark size=1.75pt, color=blue, mark=o] [unbounded coords=jump] table[x expr=\coordindex, y=planner] {figures/timeLUBM.dat};
    	\addlegendentry{Planner}	
    	\addplot [mark size=1.75pt, color=red, mark=star] [unbounded coords=jump] table[x expr=\coordindex, y=noplanner] {figures/timeLUBM.dat};
    	\addlegendentry{No Planner}
        \end{axis}
    	\end{tikzpicture}
    \pgfplotstableread[]{figures/memLUBM.dat}\datatable
	 \begin{tikzpicture}[scale=1]
	    \pgfkeys{%
	    /pgf/number format/set thousands separator = {}}
	    \begin{axis}[
    	scale only axis
    	, font=\scriptsize
    	, ylabel={\textit{\small Memory (MB)}}
    	, width=0.8\textwidth
        , legend style={at={(0.11,0.95)},anchor=north}
    	, height=0.15\textwidth
    	, ymin=10000, ymax=40000
    	, major tick length=2pt
    	, xtick=data
        , xticklabels from table={\datatable}{label}
	]
	\addplot [mark size=1.75pt, color=blue, mark=o] [unbounded coords=jump] table[x expr=\coordindex, y=planner] {figures/memLUBM.dat};
	\addlegendentry{Planner}	
	\addplot [mark size=1.75pt, color=red, mark=star] [unbounded coords=jump] table[x expr=\coordindex, y=noplanner] {figures/memLUBM.dat};
	\addlegendentry{No Planner}
    \end{axis}
	\end{tikzpicture}
\caption{Experiments on LUBM. Queries are ordered by increasing values w.r.t. the No-Planner execution.\label{fig:lubm}}
\end{figure}

\begin{figure}[h]\centering
\pgfplotstableread[]{figures/timeLUBM-LUTZ.dat}\datatable
	\begin{tikzpicture}[scale=0.99]
	\pgfkeys{%
	    /pgf/number format/set thousands separator = {}}
	\begin{axis}[
    	scale only axis
    	, font=\scriptsize
    	, x label style = {at={(axis description cs:0.5,-0.1)}}
    	, y label style = {at={(axis description cs:0.03,0.5)}}
    	, ylabel={\textit{\small Time (s)}}
    	, width=0.8\textwidth
        , legend style={at={(0.11,0.95)},anchor=north}
    	, height=0.15\textwidth
    	, ymin=200, ymax=360
    	, major tick length=2pt
    	, xtick=data
        , xticklabels from table={\datatable}{label}
	]
	\addplot [mark size=1.75pt, color=blue, mark=o] [unbounded coords=jump] table[x expr=\coordindex, y=planner] {figures/timeLUBM-LUTZ.dat};
	\addlegendentry{Planner}	
	\addplot [mark size=1.75pt, color=red, mark=star] [unbounded coords=jump] table[x expr=\coordindex, y=noplanner] {figures/timeLUBM-LUTZ.dat};
	\addlegendentry{No Planner}
    \end{axis}
	\end{tikzpicture}

\pgfplotstableread[]{figures/memLUBM-LUTZ.dat}\datatable
	\begin{tikzpicture}[scale=0.99]
	\pgfkeys{%
	    /pgf/number format/set thousands separator = {}}
	\begin{axis}[
    	scale only axis
    	, font=\scriptsize
    	, x label style = {at={(axis description cs:0.5,-0.1)}}
    	, y label style = {at={(axis description cs:0.03,0.5)}}
    	, ylabel={\textit{\small Memory (MB)}}
    	, width=0.8\textwidth
        , legend style={at={(0.11,0.95)},anchor=north}
    	, height=0.15\textwidth
    	, major tick length=2pt
    	, xtick=data
        , xticklabels from table={\datatable}{label}
	]
	\addplot [mark size=1.75pt, color=blue, mark=o] [unbounded coords=jump] table[x expr=\coordindex, y=planner] {figures/memLUBM-LUTZ.dat};
	\addlegendentry{Planner}	
	\addplot [mark size=1.75pt, color=red, mark=star] [unbounded coords=jump] table[x expr=\coordindex, y=noplanner] {figures/memLUBM-LUTZ.dat};
	\addlegendentry{No Planner}
    \end{axis}
	\end{tikzpicture}
\caption{Experiments on LUBM-LUTZ. Queries are ordered by increasing values w.r.t. the No-Planner execution.\label{fig:LUBM-LUTZ}}
\end{figure}
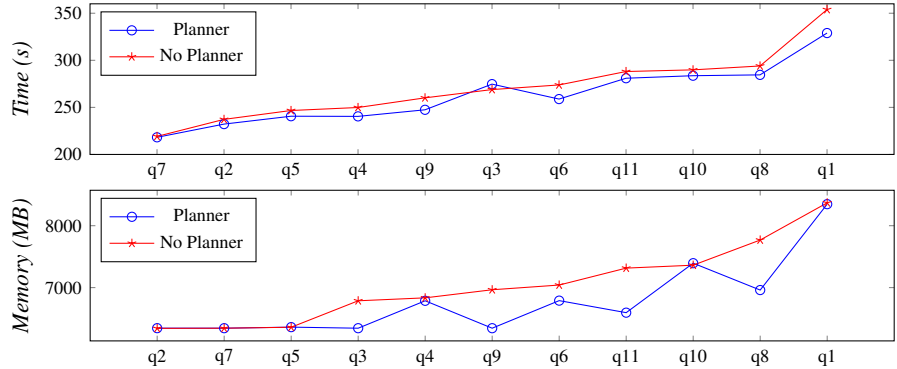
\begin{table*}[!ht]
 \begin{center}
  \includegraphics[width=\textwidth,keepaspectratio]{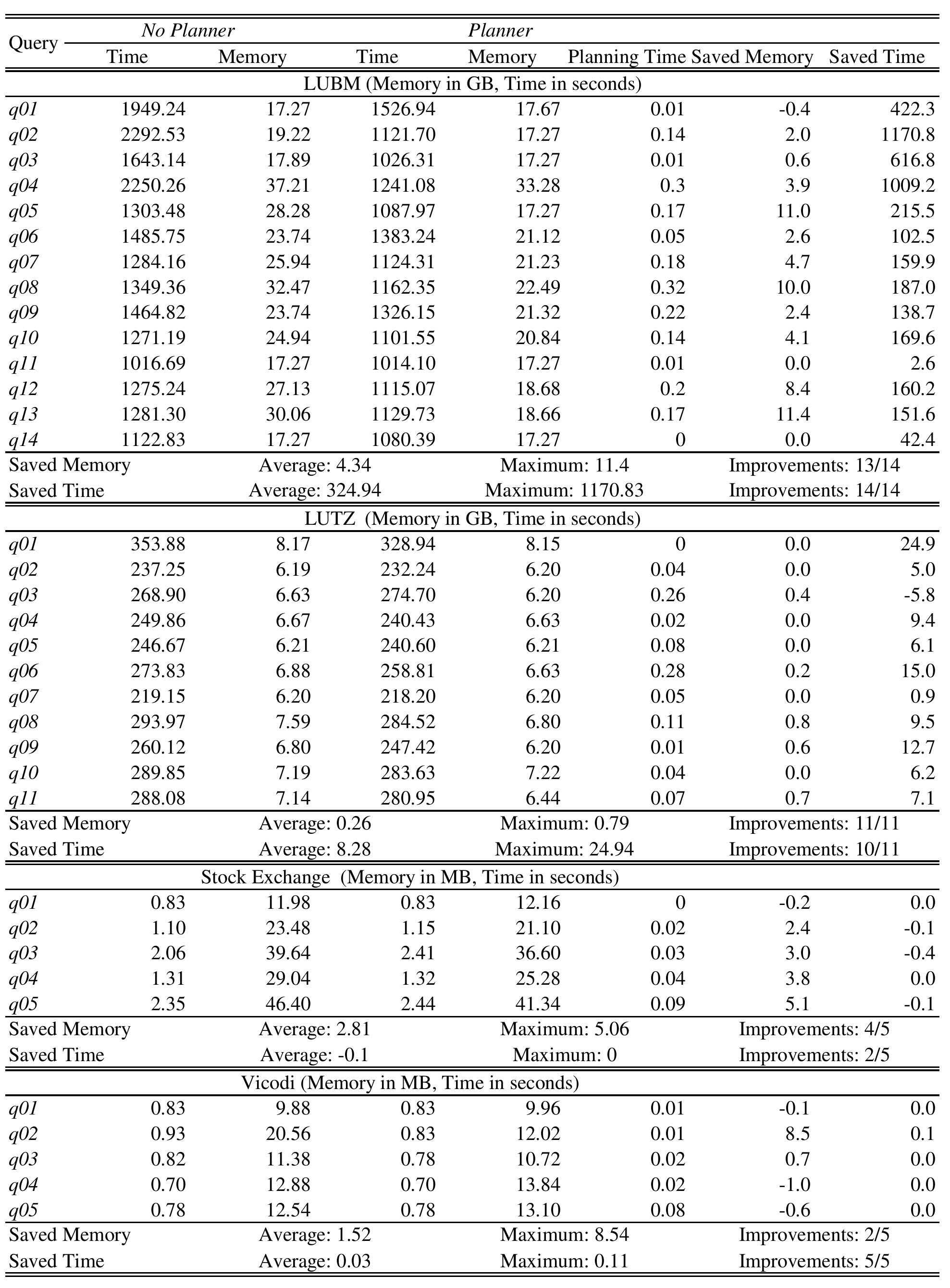}
  \caption{Experiments on LUBM, LUBM-LUTZ, Stock Exchange and Vicodi. Time is in seconds, memory is in GB for LUBM and LUBM-LUTZ and in MB for Stock Exchange and Vicodi.}
  \label{tab:all_bench}
 \end{center}
\end{table*}

\begin{table*}[!ht]
 \begin{center}
  \includegraphics[width=0.85\textwidth,keepaspectratio]{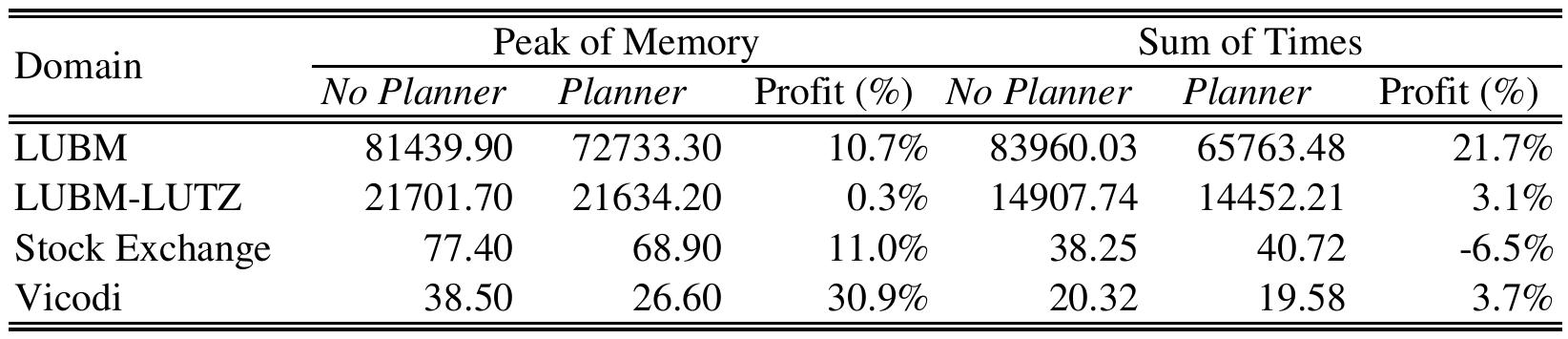}
  \caption{Statistics on LUBM, LUBM-LUTZ, Stock Exchange and Vicodi: the maximum peak of memory and the total sum of execution times computed over all databases and queries, along with the corresponding profits. Time is in seconds, memory is in MB.}
  \label{tab:all_bench2}
 \end{center}
\end{table*}

The results of our experiments are reported in Table~\ref{tab:all_bench} and in Figure~\ref{fig:lubm} and \ref{fig:LUBM-LUTZ}.

Table~\ref{tab:all_bench} shows performance in terms of average running time and memory usage of \idlv (with and without planner) computed over all considered databases per each benchmark query. Columns 2 and 3 refer to the classical computation, while columns 4 and 5 to the computation driven by the planner. In the 6th column, we reported the time spent to compute the optimal plan, in the 7th column, the memory saving per query computed as difference of the corresponding fields in columns 3 and 5. Similarly, the 8th column reports the time saving per query computed as difference of the corresponding fields in columns 2 and 4. The table reports also some aggregated data per benchmark. In particular, it shows information on the average/maximum saved memory/time, as well as the number of queries where an improvement in terms of saved memory (resp. saved time) has been obtained.  In addition, to provide a clearer picture of the behavior of the two versions of \idlv, we reported  in Figures~\ref{fig:lubm} and \ref{fig:LUBM-LUTZ} plots of the average running time and memory usage over all considered databases for the largest benchmarks: LUBM and LUBM-LUTZ.

As it can be seen, we obtained a significant saving of memory on LUBM where, for instance, the planner allows to save 11.4 GB on query \textit{q13} (about 40\% less) w.r.t. the no-planner version, and a gain both in terms of memory and time over almost all queries. Only on query \textit{q01} we experimented a small worsening on memory.
In general, no significant increase of computation time is observable and, in several cases, the planner-driven approach leads also to improvements in terms of time. This can be explained considering that indices selected by the planner, being on the overall less memory expensive, are more efficiently computable.

Concerning LUBM-LUTZ, we first note that the benefits appear less evident. This is due to the
nature of the queries in the benchmark which are constant-free and require a different (less informed) customization, as described in Section~\ref{sec:custom}. Nonetheless, the execution of \idlv driven by the planner performs generally better (both in time and memory) of the standard execution. On the queries \textit{q08}, \textit{q09} and \textit{q11} we have a memory saving of 9-10\% w.r.t. the no-planner version; moreover, we observe no worsening in memory consumption and only one case in which there is a negligible worsening in time.

As for Stock Exchange and Vicodi, although these are not data intensive domains, \idlv can benefit by the planner as well. Indeed, worsenings in terms of memory range from 1\% to 7\% in a few queries which are somehow expected when measuring memory of the order of megabytes.

Further aggregated data and statistics on the results are given in Table~\ref{tab:all_bench2}. This shows, for both the tested versions of \idlv and for each benchmark, the maximum peak of memory and the total sum of execution times computed over all databases and queries, along with the corresponding profits. In all benchmarks, the peak of memory when the planner is used is less than the one obtained using the standard version of \idlv.   Regarding times, although we experimented a small worsening for Stock Exchange (6.5\%), we observe a general improvement which is greater than 20\% in our large-scale benchmark.

\section{Conclusion}
In this work we introduced an evaluation planner for Datalog programs. The planner has been conceived to be applied to ontology-based query answering contexts, where often, in case of large databases, standard approaches are not convenient/applicable due to memory consumption. It relies on an ASP program that computes the plan, intended as an indexing schema for the database together with a body-ordering for each rule in the program. The computed plan minimizes the overall cost (in term of memory consumption) of indices; moreover, the usage of the plan with the \dlv system allows to further reduce memory usage since some expensive internal optimizations of \dlv can be disabled. Results of the experiments conducted on popular ontological benchmarks confirm the effectiveness of the approach.
Eventually, precomputing offline an evaluation plan plays an extremely important role in reasoners with a server-like behavior, since this allows for further reducing the time required by the actual computation.

\section*{Acknowledgments}
This work has been partially supported by MIUR under project ``Declarative Reasoning over Streams'' (CUP H24I17000080001) -- PRIN 2017, by MISE under project ``S2BDW'' (F/050389/01-03/X32) -- ``Horizon2020'' PON I\&C2014-20, by Regione Calabria under project ``DLV LargeScale'' (CUP J28C17000220006) -- POR Calabria 2014-20.




\bibliographystyle{acmtrans}
\newcommand{\SortNoOp}[1]{}


\end{document}